%% file: MTE_V_E.TEX
\documentstyle[12pt,mkstyle,twoside]{article}
\title{Mathematical Theory of Evidence Versus Evidence}

\newcommand{\rDef}[1]{Def.\ref{#1}}
\newcommand{\Bem}[1]{}
\newcommand{\Bemerkung}[1]{}

\newcommand{\rTab}[1]{Table \ref{#1}} 

\font\gh=eufm10 scaled \magstep1

\newcommand{\Prob}[2]{{ {\mbox{\gh Prob} ^{#2(#1)}} 
                         \atop {_{#1}} 
                     }}
\date{}
\begin{document}

\machetitel

\begin{abstract}
This paper is concerned with the apparent greatest weakness of the
Mathematical Theory of Evidence (MTE) of Shafer \cite{Shafer:76}, which has
been strongly criticized by Wasserman \cite{Wasserman:92ijar}. \\
Weaknesses of Shafer's proposal \cite{Shafer:90b} of probabilistic
interpretation of MTE belief functions is demonstrated. Thereafter a new
probabilistic interpretation  of MTE conforming both to definition of belief
function and to Dempster's rule of combination of independent evidence. It is
shown that shaferian conditioning of belief functions on observations
\cite{Shafer:90b}  may be treated as selection combined with modification of
data, that is data is not viewed as it is but it is casted into one's beliefs
in what it should be like..\\
\end{abstract}
 
\section{Introduction}

Wasserman in    \cite{Wasserman:92ijar}  raised  serious  concerns 
against
the Mathematical Theory of Evidence (MTE) developed by Dempster and Shafer
since 1967 (see 
\cite{Shafer:90ijar} for a thorough review of this theory). 
One of arguments against MTE is related to  Shafer's attitude towards
frequencies. 
Shafer in  \cite{Shafer:90ijar} claims that probability theory developed over
last years from the old-style frequencies towards modern subjective
probability theory within the framework of bayesian theory. By analogy he
claims that the very attempt to consider relation between MTE and frequencies
is old-fashioned and out of date and should be at least forbidden - for the
sake of progress of humanity. 
Wasserman opposes this view (\cite{Wasserman:92ijar}, p.371)
reminding  "major
 success story in Bayesian theory",  the exchangeability theory of 
de Finetti \cite{deFinetti:64}. It treats frequencies as special case of
bayesian belief.  "The Bayesian 
theory contains within it a definition of frequency probability and a 
description of the exact assumptions necessary to invoke that 
definition" \cite{Wasserman:92ijar}. Wasserman dismisses Shafer's suggestion
that probability relies on analogy of frequency. .%\\

Shafer, on the other hand, lets frequencies live a separate life. MTE beliefs
and frequencies are separated. But in this way we are left without a
definition of frequentistic belief function \cite{Wasserman:92ijar}. %\\

This paper presents an attempt to introduce a frequentistic definition of 
belief function which shall be fully compatible with MTE. Section 2 reminds
briefly basic definitions of MTE. Section 3 demonstrates shortcomings of
Shafer's view of frequencies for belief functions. Section 4 presents our
denotation. Section 5 introduces the new frequentistic interpretation of MTE.
Section 6 summarizes the new interpretation. %\\

\section{Formal  Definitions of MTE}

Let $\Xi$ be a finite  set of elements called elementary events. 
Any subset of $\Xi$ be a composite event. $\Xi$ be called also the 
frame of discernment.\\
A basic probability assignment function m:$2^\Xi  \rightarrow [0,1]$
such that  $$  \sum_{A \in 2^\Xi } |m(A)|=1 $$
$$  m(  \emptyset) =0 $$
$$\forall_{A \in 2^\Xi} \quad  0 \leq  \sum_{A \subseteq B} m(B)$$
%\\
($|.|$ - absolute value.\\
      
A belief function be defined as Bel:$2^\Xi \rightarrow [0,1]$ so that 
 $Bel(A) = \sum_{B \subseteq A} m(B)$
A plausibility function be Pl:$2^\Xi \rightarrow [ 0,1]$  with 
$\forall_{A \in 2^\Xi} \  Pl(A) = 1-Bel(\Xi-A )$
A commonalty function be Q:$2^\Xi \rightarrow [0,1]$ with 
 $\forall_{A \in 2^\Xi} \quad Q(A) = \sum_{A \subseteq B} m(B)$

Furthermore, a Rule of Combination of two Independent Belief Functions 
$Bel_1$,
 $Bel_2$ Over the Same Frame of Discernment (the so-called Dempster-Rule),
denoted 
    $$Bel_{E_1,E_2}=Bel_{E_1} \oplus Bel_{E_2}$$ 
 is defined as follows: :
$$m_{E_1,E_2}(A)=c \cdot  \sum_{B,C; A= B \cap C} m_{E_1}(B) \cdot 
m_{E_2}(C)$$ (c - constant normalizing the sum of $|m|$ to 1)%\\

 Whenever $m(A) > 0$, we say that A is the focal  point  of  the 
Bel-Function.
 If the only focal point of a belief function is $\Xi$ ($m(\Xi)=1$), then 
Bel is called vacuous belief function (it does not contain any information on
whatever value is taken by the variable).

 Let B be a subset of $\Xi$, called 
evidence,
 $m_B$ be a basic probability assignment such that $m_B(B)=1$ and $m_B(A)=0$
for any A different from B. Then the conditional belief function $Bel(.||B)$
representing the belief function $Bel$ conditioned on evidence  B 
is defined
as: $Bel(.||B)=Bel \oplus Bel_B$  (Compare \cite{Shafer:90b}) %\\

\section{Basic Problems with Frequencies in MTE}

Shafer in  \cite{Shafer:90}, \cite{Shafer:90b} gave the following formal
probabilistic interpretation of belief function: 
Let $Pr$ be a probabilistic measure over the sample space  $\Omega$, 
let   $\Gamma$ be a function  $\Gamma:\Omega \rightarrow 2^\Xi$. Then  $Bel$ 
over the space (frame of discernment) $\Xi$ is given as:%\\
$$Bel(A) = Pr(\{\omega \in \Omega | \Gamma(\omega) \subseteq A \})$$ %\\
Then clearly      
$$m  (A) = Pr(\{\omega \in \Omega | \Gamma(\omega)  =        A \})$$ %\\
and  
$$Pl (A) = Pr(\{\omega \in \Omega | \Gamma(\omega) \cap A 
\ne \emptyset\})$$ %\\

Let us consider the database in \rTab{teins}.

\begin{table}
\caption{Example of $\Gamma$ function}
\label{teins}
\begin{tabular}{r|r|r|r}
No.  & A & D & $\Gamma$ \\
\hline
1 & $a_1$ & $d_1$ & $\{d_1     \}$ \\
2 & $a_2$ & $d_2$ & $\{d_2,d_3 \}$ \\
3 & $a_2$ & $d_3$ & $\{d_2,d_3 \}$ \\
4 & $a_3$ & $d_3$ & $\{d_3     \}$ \\
5 & $a_4$ & $d_1$ & $\{d_1     \}$ \\
\end{tabular}
\end{table}
%\\
Let the measurable A take values  $a_1,a_2,a_3,a_4$, and let the
non-observable attribute D take values  $d_1,d_2,d_3$.
Let us define the function $\Gamma$ as capability to predict values of
attribute D given A and let us calculate it based on training sample contained
in  \rTab{teins}. We see that if A takes value 
 $a_1$, then we know that D takes value  $d_1$ - hence 
$\Gamma(A=a_1)=\{d_1\}$. Similarly values $a_3$ and  $a_4$ of attribute A 
determine uniquely the value of attribute D. But in case of 
$A=a_2$ we have an ambiguity: D is equal either  $d_2$ or  $d_3$.
Hence  $\Gamma(A=a_1)=\{d_2,d_3\}$. Now, assuming frequency probabilities
from  \rTab{teins} we calculate easily from Shafer's formula:\\
\begin{tabular}{ccc}
$m(\{d_1\})=0.4  $     &    $Bel(\{d_1\})=0.4$ &  $Pl(\{d_1\})=0.4$  \\
$m(\{d_2\})=0    $     &    $Bel(\{d_2\})=0  $ &  $Pl(\{d_2\})=0.4$  \\
$m(\{d_3\})=0.2  $     &    $Bel(\{d_3\})=0.2$ & $Pl(\{d_3\})=0.6$  \\
$m(\{d_1,d_2\})=0    $ & $Bel(\{d_1,d_2    \})=0.4$&$Pl(\{d_1,d_2  \})=0.8$\\
$m(\{d_1,d_3\})=0    $ & $Bel(\{d_1,    d_3\})=0.6$&$Pl(\{d_1,    d_3\})=1 $\\
$m(\{d_2,d_3\})=0.4  $ & $Bel(\{    d_2,d_3\})=0.6$&$Pl(\{   d_2,d_3\})=0.6$\\
$m(\{d_1,d_2,d_3\})=0$ & $Bel(\{d_1,d_2,d_3\})=1  $&$Pl(\{d_1,d_2,d_3\})=1 $\\
\end{tabular}%\\

In probability theory two variables are independent if 
$Pr(A \cap B)=Pr(A) \cdot Pr(B)$. Let us consider two measurables A and B 
from \rTab{tzwei}

\begin{table}
\caption{Example of $\Gamma'$ function for a variable  B
independent of A} \label{tzwei}
\begin{tabular}{r|r|r|r|r|r}
No.  & A & B & D & $\Gamma$ & $\Gamma'$ \\
\hline
 1 & $a_1$ & $b_1$ & $d_1$ & $\{d_1     \}$ & $\{d_1,d_2,d_3 \}$ \\
 2 & $a_2$ & $b_1$ & $d_2$ & $\{d_2,d_3 \}$ & $\{d_1,d_2,d_3 \}$ \\
 3 & $a_2$ & $b_1$ & $d_3$ & $\{d_2,d_3 \}$ & $\{d_1,d_2,d_3 \}$ \\
 4 & $a_3$ & $b_1$ & $d_3$ & $\{d_3     \}$ & $\{d_1,d_2,d_3 \}$ \\
 5 & $a_4$ & $b_1$ & $d_1$ & $\{d_1     \}$ & $\{d_1,d_2,d_3 \}$ \\
 6 & $a_1$ & $b_2$ & $d_1$ & $\{d_1     \}$ & $\{d_1,d_2,d_3 \}$ \\
 7 & $a_2$ & $b_2$ & $d_2$ & $\{d_2,d_3 \}$ & $\{d_1,d_2,d_3 \}$ \\
 8 & $a_2$ & $b_2$ & $d_3$ & $\{d_2,d_3 \}$ & $\{d_1,d_2,d_3 \}$ \\
 9 & $a_3$ & $b_2$ & $d_3$ & $\{d_3     \}$ & $\{d_1,d_2,d_3 \}$ \\
10 & $a_4$ & $b_2$ & $d_1$ & $\{d_1     \}$ & $\{d_1,d_2,d_3 \}$ \\
11 & $a_1$ & $b_3$ & $d_1$ & $\{d_1     \}$ & $\{d_1,d_2,d_3 \}$ \\
12 & $a_2$ & $b_3$ & $d_2$ & $\{d_2,d_3 \}$ & $\{d_1,d_2,d_3 \}$ \\
13 & $a_2$ & $b_3$ & $d_3$ & $\{d_2,d_3 \}$ & $\{d_1,d_2,d_3 \}$ \\
14 & $a_3$ & $b_3$ & $d_3$ & $\{d_3     \}$ & $\{d_1,d_2,d_3 \}$ \\
15 & $a_4$ & $b_3$ & $d_1$ & $\{d_1     \}$ & $\{d_1,d_2,d_3 \}$ \\
16 & $a_1$ & $b_4$ & $d_1$ & $\{d_1     \}$ & $\{d_1,d_2,d_3 \}$ \\
17 & $a_2$ & $b_4$ & $d_2$ & $\{d_2,d_3 \}$ & $\{d_1,d_2,d_3 \}$ \\
18 & $a_2$ & $b_4$ & $d_3$ & $\{d_2,d_3 \}$ & $\{d_1,d_2,d_3 \}$ \\
19 & $a_3$ & $b_4$ & $d_3$ & $\{d_3     \}$ & $\{d_1,d_2,d_3 \}$ \\
20 & $a_4$ & $b_4$ & $d_1$ & $\{d_1     \}$ & $\{d_1,d_2,d_3 \}$ \\
\end{tabular}
\end{table}

Function $\Gamma$ be, as previously, be prediction of value of variable D
based on value of A, and  $\Gamma'$ be prediction of variable D given value of
 B. Let us define %\\
$$Bel(Z) = Pr(\{\omega \in \Omega | \Gamma(\omega) \subseteq Z \})$$ %\\
$$Bel'(Z) = Pr(\{\omega \in \Omega | \Gamma'(\omega) \subseteq Z \})$$ %\\

Let us imagine that we want to combine information from attributes A and B to
improve prediction of D by formulating a new function  $\Gamma"$ being the
base for a new belief function:%\\
$$Bel"(Z) = Pr(\{\omega \in \Omega | \Gamma"(\omega) \subseteq Z \})$$ %\\

As observations being basis of functions  $\Gamma$  and 
 $\Gamma'$ are obviously independent, so one would expect that the belief
function  $Bel"$ is simply the combination OF INDEPENDENT EVIDENCE 
$Bel$  and  $Bel'$ via Dempster rule.  And this is in fact the case:%\\
$$Bel"=Bel \oplus Bel'$$ %\\
But there is one weak point in all of this: $Bel'$ is (and will always be) a
vacuous belief function, hence it does not contribute anything to our
knowledge of the value of the attribute. Reverting this example we can say
that whenever we combine two non-vacuous belief functions, then the
measurements underlying their empirical calculation are for sure statistically
dependent. So we claim that:\\
{\em Under Shafer's frequentist interpretation, if two belief functions are
(statistically) independent then at least one of them is non-informative.} %\\

Another practical limitation of Shafer's probabilistic interpretation is 
consideration of conditional beliefs. %\\
Let as look at \rTab{tdrei}. 

\begin{table}
\caption{Scheme of creation of conditional probability}
\label{tdrei}
\begin{tabular}{r|r|r|r}
No.  & A & $A=a_1 \lor A=a_2$ & $A|A=a_1 \lor A=a_2$ \\
\hline
1 & $a_1$ & yes& $a_1$ \\
2 & $a_2$ & yes& $a_2$ \\
3 & $a_2$ & yes& $a_2$ \\
4 & $a_3$ & no & $ - $ \\
5 & $a_4$ & no & $ - $  \\
\end{tabular}
\end{table}

If we want to calculate conditional probability of  $A=a_1$ given         
observation that A takes only one of values $a_1$ or  $a_2$, 
we select cases from the database fitting the condition   $A=a_1 
\lor A=a_2$, and thereafter within this subset we calculate frequency
probabilities:  $Pr(A=a_1|A=a_1 \lor A=a_2)=1/3=0.33$.%\\
Now, based on \rTab{tvier} let us run similar procedure for MTE beliefs.%\\

\begin{table}
\caption{Example of conditioning in MTE}
\label{tvier}
\begin{tabular}{r|r|r|r|r|r}
No.  & A & D & $\Gamma$ & $\Gamma \cap\{d_1,d_2 \}$ & $\Gamma'=$\\
     &   &   &          & $\neq \emptyset$   & $\Gamma \cap\{d_1,d_2 \}$\\
\hline
1 & $a_1$ & $d_1$ & $\{d_1     \}$ & yes & $\{d_1     \}$ \\
2 & $a_2$ & $d_2$ & $\{d_2,d_3 \}$ & yes & $\{d_2     \}$  \\
3 & $a_2$ & $d_3$ & $\{d_2,d_3 \}$ & yes & $\{d_2     \}$  \\
4 & $a_3$ & $d_3$ & $\{d_3     \}$ & no  & $      -     $  \\
5 & $a_4$ & $d_1$ & $\{d_1     \}$ & no  & $      -     $   \\
\end{tabular}
\end{table}

Let us assume that we want to find out our degree of belief in values of D
given that only values  $d_1$ or $d_2$ are allowed.  For this purpose
we restrict the set of cases to those cases  $\Omega'$ for which our function
 $\Gamma$ has non-empty intersection with the set of values of interest. 
For this group of cases we define the function 
 $\Gamma'(\omega)= \Gamma(\omega) \cap \{d_1,d_2
\}$. Let :\\
$$Bel(Z) = Pr(\{\omega \in \Omega | \Gamma(\omega) \subseteq Z \})$$ %\\
$$Bel'(Z) = Pr(\{\omega \in \Omega' | \Gamma'(\omega) \subseteq Z \})$$ %\\

Additionally let us define the simple support function $Bel"$ such that
$m"(\{d_1,d_2\})=1$. It is easily seen that:\\
$$Bel' = Bel \oplus Bel"$$%\\
(as expected because the expression  $Bel \oplus Bel"$%\\
means shaferian conditioning on event
$\{d_1,d_2\}$). 
And everything would be O.K. if it were not that the function 
$\Gamma'$ has little to do with the non-observable attribute $D$ - compare
line no.3 of \rTab{tvier}. 
Let us remind that function $\Gamma$ 
represented by definition for a given observed value of variable A
the set of potentially possible values of attribute D, deducible from the
training sample. For every object $\omega$, if we know the true value 
$a$ of $A$ one of the values from the set $\Gamma(\omega)$ was the true value
of
D for this object $\omega$. But within  $\Gamma'(\omega)$ the true value of
attribute D does not need to be contained - compare
line no.3 of \rTab{tvier}. 
But, let us remind, Shafer claimed 
\cite{Shafer:90,Shafer:90b} that function  $\Gamma$ indicates that
the variable takes for object 
 $\omega$ one of the values  $\Gamma(\omega)$. But we have just demonstrated
that already after a single step of conditioning function  $\Gamma'$
simply tells lies. Its meaning is not dependent solely on subpopulation
$\Omega'$, to which it refers, but also on the history, how this population
was selected. But we had for probability distributions that after conditioning
a variable for not rejected objects took always those values which were
indicated by the result of conditioning.%\\

Both above failures of Shafer's probabilistic interpretation of his own
theory of evidence were driving forces behind the elaboration of a new
probabilistic interpretation of MTE presented subsequently. \\

We shall summarize this section saying that: {\em Shafer's probabilistic
interpretation of Dempster's \& Shafer's Mathematical Theory of Evidence is
not compatible with this theory: It does not fit the Dempster's rule of
combination of independent evidence}. .%\\
%\\

\section{Denotation}

F. Bacchus in his paper \cite{Bacchus:90} on axiomatization of 
probability theory and 
first order logic shows that probability should be considered as a quantifier 
binding free variables
in first order logic expressions just like universal and existential 
quantifiers do. So if e.g. $\alpha(x)$ is an open expression with a free 
variable $x$ then   $[\alpha(x)]_x$ means the probability of truth of the 
expression  $\alpha(x)$. 
(The quantifier $[]_x$ binds the free variable $x$ and yields a numerical 
value ranging from 0 to 1 and meeting all the Kolmogoroff axioms). 
Within the expression  $[\alpha(x)]_x$ the variable 
$x$ is bound. See \cite{Bacchus:90} on justification why other types of 
 integration of probability theory and first order logic or propositional 
logic fail. Also for justification of rejection
of the traditional view of probability as a function over sets.
While sharing Bacchus' view, we find his notation a 
bit cumbersome so we change it to be similar to the universal and 
existential quantifiers throughout this paper.
Furthermore, Morgan \cite{Morgan:91} insisted that the probabilities be 
always considered in close connection with the population they refer to.  
 Bacchus' expression 
$[\alpha(x)]_x$ we rewrite as:\\
  $\Prob{x}{P}\alpha(x)$ - the probability of  $\alpha(x)]$ being true within 
the population P. The  P (population) is a unary predicate with P(x)=TRUE 
indicating that the object x($\in \Omega$, that is element of a universe of 
objects) belongs to the population under considerations. If P and P' are 
populations such that $\forall_x P'(x)\rightarrow P(x)$ (that is membership 
in P' implies membership in P, or in other words: P' is a subpopulation of 
P), then we distinguish two cases:\\
case 1: $(\Prob{x}{P}P'(x))=0$ (that is probability of membership in P' with 
respect to P is equal 0) - then (according to \cite{Morgan:91}) for any 
expression $\alpha(x)$ in free variable x the following holds for the 
population P': $(\Prob{x}{P'}\alpha(x))=1$\\
case 2: $(\Prob{x}{P}P'(x))>0$then (according to \cite{Morgan:91} for any 
expression $\alpha(x)$ in free variable x the following holds for the 
population P': 
$$(\Prob{x}{P'}\alpha(x))=  \frac {\Prob{x}{P}(\alpha(x) \land P'(x))}
                                {\Prob{x}{P}P'(x)}$$  %\\
We also use the following (now traditional) mathematical symbols:\\
$\forall_{x}\alpha(x)$ - always  $\alpha(x)$ (universal quantifier) \\
$\exists_{x}\alpha(x)$ - there exists an x such that $\alpha(x)$ 
(existential quantifier) \\
\begin{tabular}{lp{11cm}}%\\
$\alpha \land \beta$ & - logical AND of expressions\\
$\bigwedge_{B} \alpha(B)$  & - logical AND over all  instantiations of
the expression $\alpha(B)$ in free 
variable $B$\\
$\alpha \lor \beta$  & - logical OR of expressions\\
$\bigvee_{B} \alpha(B)$  & - logical OR over all  instantiations of
the expression $\alpha(B)$ in free 
variable $B$\\
$\lnot$  & - logical negation\\
$P \cap Q$  & - intersection of two sets\\
$P \cup Q$  & - union of two sets\\
\end{tabular}%\\

\section{A New Interpretation of Belief Functions}

The empirical meaning of a
new interpretation of the MTE Belief function will be explained
by means of the following example:

\begin{Bsp} \label{COOTexample}
Let us consider a daily-life example. Buying a bottle of hair shampoo is 
not a trivial task from both the side of the consumer and the 
manufacturer. If the consumer arrives at the consciousness that the shampoos 
may fall into one of the four categories: 
high quality products 
(excellent for maintaining cleanness and health of the consumer)
 (H), 
moderate quality products (keeping just all Polish industry standards) (M), 
suspicious products  (violating some industry standards) (S)
 and products dangerous for health and life (containing bacteria or fungi or 
other microbes causing infectious or invasive diseases, containing 
cancerogenous or poisonous substances etc.) (D), 
he has a hard time upon leaving his house for shopping. Clearly, precise 
chemical, biochemical and medical tests exist which may precisely place
the product into one of those obviously exclusive categories. But the
 Citizen\footnote{The term "Citizen" was a fine socialist time descriptor 
allowing to avoid the cumbersome usage of words like "Mr.", "Mrs."  and 
 "Miss"} Coot\footnote{This family name was coined as abbreviation for 
"Citizen 
Of Our Town"} usually neither has a private chemical laboratory nor enough 
money to make use of required services. Hence Citizen Coot coins a personal 
set of "quality" tests $M ^1$ mapping the pair (bottle of shampoo, quality) 
into the set \{TRUE, FALSE\} (the letter O - object - stands for bottle of 
shampoo, H, M, S, D indicate quality classes: high, moderate, suspicious, 
dangerous): \\
\begin{enumerate}%\\
\item If the shampoo is heavily advertised on TV then it is of 
high quality ($M ^1(O,\{H\})=TRUE$) and otherwise not ($M ^1(O,\{H\})=FALSE$).
%\\
\item If the name of the shampoo was never heard on TV, 
 but the 
bottle looks fine 
(pretty colors, aesthetic shape of the bottle), then the shampoo must be of 
moderate quality  ($M ^1(O,\{M\})=TRUE$)  and otherwise not 
($M ^1(O,\{M\})=FALSE$).%\\
\item   If  the packaging is not fine or  the date of production is not 
readable on the bottle or 
the product is out of date, but the shampoo  smells acceptably 
otherwise 
 then it is suspicious  ($M ^1(O,\{S\})=TRUE$)  and 
otherwise not ($M ^1(O,\{S\})=FALSE$).%\\
\item  If either the packaging is not fine or 
the date of production is not readable on the bottle or 
the product is out of date,
and at the same time 
the shampoo smells awfully,  then it is dangerous  
($M ^1(O,\{D\})=TRUE$  and otherwise not ($M ^1(O,\{D\})=FALSE$).%\\
\end{enumerate} 

Notice that the criteria are partially rational: a not fine looking bottle 
may in fact indicate some decaying processing of the shampoo or at least that 
the product remains for a longer time on the shelf already. Bad smell is 
usually caused by development of some bacteria dangerous for human health.%\\
Notice also that test for high and moderate quality are enthusiastic, while 
the other two are more cautious. %\\

Notice that the two latter tests are more difficult to carry out in a shop
than the leading two (the shop assistant would hardly allow to open a  
bottle before buying). 
Also, there may be no time to check whether the shampoo was actually
advertised on TV or not (as the son who carefully watches 
all the running advertisements stayed home and does his lessons).
Hence some simplified tests may be quite helpful:
\begin{itemize} 
\item $M ^1(O,\{S,D\})$: If  the packaging is not fine or the product is out 
of date or the production date is not readable  then the product is 
either suspicious or dangerous  ($M ^1(O,\{S,D\})=TRUE$  and otherwise not 
($M ^1(O,\{D,S\})=FALSE$).
 .%\\
\item $M ^1(O,\{H,M\})$: If the packaging looks fine, then the product is 
either of high    or moderate quality  ($M ^1(O,\{M,H\})=TRUE$  and otherwise 
not ($M ^1(O,\{M,H\})=FALSE$)..%\\
\end{itemize}
 Clearly these  tests are far from being precise ones, but for the Citizen 
Coot 
no better tests will be ever available. What is more, they are not exclusive: 
if one visits a dubious shop at a later hour, one may buy a product meeting 
both  $M ^1(O,\{H\})$ and  $M ^1(O,\{D\})$ as defined above ! %\\

Let us assume we have two types of shops in our town: good ones (G) and bad 
ones (B). (Let $M ^2:\Omega \times 2^\{G,B\} \rightarrow \{TRUE, FALSE\}$ 
indicate for each shampoo in which shop type it was available. Further, let  
$M ^3:\Omega \times 2^{\{ H, M, S, D \} \times \{G,B\}} \rightarrow \{TRUE, 
FALSE\}$ indicate for each shampoo both its quality and the type of shop it 
was available from. Let clearly $M ^1(O,Quality) \land M ^2(O,Shop)=M 
^3(O,Quality \times Shop)$.\\
The good shops are those  with  new  furniture,  well-clothed  shop 
assistants. Bad 
ones are those with always dirty floor or old furniture, or badly clothed 
shop assistants. Clearly, again, both shop categories may be considered 
(nearly) exclusive as seldom well clothed shop assistants 
 do not care of floors. Let us assume we have obtained the 
statistics of shampoo sales in our town presented in Table \ref{statshamp}.%\\

\begin{table} \label{statshamp}
\caption{Sold shampoos statistics}
\begin{center}
\begin{tabular}{r|rrr|r}
      Quality true for & Shop type  B &  G  & B,G & Total\\
\hline%\\
          H &     20   & 100 &  70 & 190  \\
          M &     80   & 100 & 110 & 290      \\
          S &     50   &   5 &  15 &  70      \\
          D &     10   &   1 &   3 &  14     \\
        H,S &     15   &  10 &  14 &  39      \\
        M,S &     30   &  20 &  25 &  75      \\
        H,D &      8   &   2 &   3 &  13     \\
        M,D &     15   &   7 &  10 &  32      \\
\hline%\\
      total &    228   & 245 & 250 & 723      \\
\end{tabular}
\end{center}
\end{table}
%\\
%\vspace{0.5cm}
%
Rows and columns are marked with those singleton tests which were passed
(e.g. in the left upper corner there are 20 
shampoo bottles sold in an undoubtedly bad shop and having exclusively
high quality, that is for all those bottles (O)
$M ^1(O,\{H\})=TRUE$, $M ^1(O,\{M\})=FALSE$,  $M ^1(O,\{S\})=FALSE$, $M 
^1(O,\{D\})=FALSE$, and  $M ^2(O,\{B\})=TRUE$,  $M ^2(O,\{G\})=FALSE$.)
The measurement of $M ^1(O,\{H\})$ would yield TRUE for 190+39+13  =242 
bottles and 
FALSE for the remaining 581 bottles, the measurement of $M ^1(O,\{D\})$ would 
yield 
TRUE for 14+13+32=59 bottles, and FALSE for the remaining 664 bottles. The 
 measurement $M ^1(O,\{S,D\})$ will turn true in 70+14+ 39+75+ 13+12 =343 
cases and FALSE in the remaining 480 cases.%\\
\end{Bsp}
%\\

In general let
us assume that we know that objects of a  population  can 
be described by an  intrinsic attribute  X taking 
exclusively   one   of   the   n   discrete   values   from   its 
domain $\Xi=\{v_1,v_2,...,v_n\}$ . Let us  assume  furthermore 
that to obtain knowledge of the actual value taken by  an  object 
we must apply a measurement method (a system of tests) $M$  

\begin{df} \label{MDef}
$X$ be a set-valued attribute taking as its values non-empty
subsets of a finite domain $\Xi$.
By a measurement method 
of value of the attribute $X$
we understand a function:
 $$M: \Omega \times 2^\Xi \rightarrow \{TRUE,FALSE\}$$. 
where $\Omega$ is the set of objects, (or population of objects)
such that 
\begin{itemize}
\item
 $ \forall_{\omega; \omega \in \Omega} \quad
M(\omega,\Xi)=TRUE$ (X takes at least one of values from $\Xi$)
\item
 $ \forall_{\omega; \omega \in \Omega} \quad
M(\omega,\emptyset)=FALSE$ 
\item 
 whenever 
$M(\omega,A)=TRUE$
for $\omega \in \Omega$, $A \subseteq \Xi$
 then for any $B$ such that $A \subset B$ $M(\omega,B)=TRUE$   
holds,
 \item 
 whenever 
$M(\omega,A)=TRUE$
for $\omega \in \Omega$, $A \subseteq \Xi$ and if $card(A)>1$ then there 
exists  $B$, $B \subset A$ such that $M(\omega,B)=TRUE$ holds.
\item 
for every $\omega$ and every $A$
either  
$M(\omega,A)=TRUE$  or 
 $M(\omega,A)=FALSE$ (but never both).
 \end{itemize}
$M(\omega,A)$  tells us whether or not any of the elements of the set A 
belong to the actual value of the attribute $X$ for the object $\omega$.%\\
 \end{df}

The measuring function M(O,A), if it takes the value TRUE,  
states for an object O and a set A of values from the domain of X  
that  the X 
takes for this object (at least) one of the  values  in A. %\\

     Let us furthermore assume that with each application of  the 
measurement  procedure  some  costs  are  connected,   increasing 
roughly with the decreasing size of the tested set A so that  we 
are ready to accept results of previous measurements in the  form 
of pre-labeling of the population. So 

\begin{df}
A {\em label} $L$ of an object $\omega \in \Omega$ is a subset of the domain
$\Xi$ of the attribute $X$. \\
A {\em labeling}  under the measurement method $M$  is a function $l: \Omega 
\rightarrow 2^\Xi$ such that for any object  $\omega \in \Omega$ either
$l(\omega)=\emptyset$ or $M(\omega,l(\omega))=TRUE$.\\
Each {\em labelled object}  (under the labeling $l$) 
consists of a 
pair $(O_j,L_j)$, $O_j$ - the j$^{th}$ object, $L_j=l(O_j)$ - its label.\\
By a {\em population  under the labeling $l$} we understand the predicate 
$P:\Omega \rightarrow \{TRUE,FALSE\}$ of the form 
$P(\omega)=TRUE \  iff \ l(\omega) \neq \emptyset$
(or alternatively, the set of objects  for which this predicate is true) \\
 If for every  object of the 
population the label is equal 
 to $\Xi$ then  we  talk  of  an  {\em unlabeled  population} (under the 
labeling $l$), otherwise of a {\em pre-labelled} one.
\end{df}

     Let  us  assume  that  in  practice  we  apply  a   modified 
measurement method 
$M_l$ being a function:

\begin{df} 
Let $l$ be a labeling under the measurement method $M$. 
Let us consider the population under this labeling.
The modified measurement method 
$$M_l:
 \Omega \times 2^\Xi \rightarrow 
\{TRUE,FALSE\}$$
where $\Omega$ is the set of objects, 
is is defined as  
$$M_l(\omega,A)= M(\omega,A \cap l(\omega) )$$  
(Notice that 
$M_l(\omega,A)=FALSE$ whenever $A \cap l(\omega)= \emptyset$.)
\end{df}

For a labeled object $(O_j,L_j)$  ($O_j$ - proper object, 
$L_j$  - 
its label)  and a set A of values from the domain of X, 
the modified measurement method tells us 
that $X$ takes one of the values in A if and only if it takes in fact 
a value from intersection of A and $L_j$.
 Expressed   differently,   we 
discard a priori any attribute not in the label.%\\

Please pay attention also to the fact, that given a population P for which 
the measurement method $M$ is defined, the labeling $l$ (according to its 
definition) selects a subset of this population, possibly a proper subset, 
namely the population P'
under this labeling. 
$P'(\omega)=P(\omega) \land M(\omega,l(\omega))$. 
Hence also $M_l$ is defined possibly for the "smaller" 
population P' than $M$ is. \\

\begin{Bsp} To continue Citizen Coot example, we may
believe that in good shops only moderate and high quality products
are available, that is we assign to every shampoo $\omega$
the label $l(\omega)=\emptyset$
(we discard it from our register)
 if $\omega$ denies our belief
that there are no suspicious nor dangerous products in a good shop,
and $l(\omega)=\{H,M\}$ if it is moderate or high quality
product in a good shop
and
$l(\omega)=\Xi$  to all the other products. After this 
rejection of shampoos not fitting our beliefs we have to do 
with (a bit smaller) sold-shampoos-population from Table 
\ref{modstatshamp}.
%\\

\begin{table} \label{modstatshamp}
\caption{Modified sold shampoos statistics}
\begin{center}
\begin{tabular}{r|rrr|r}
      Quality true for & Shop type  B &  G  & B,G & Total\\
\hline%\\
          H &     20   & 112 &  70 & 202  \\
          M &     80   & 127 & 110 & 317      \\
          S &     65   &   0 &   0 &  65      \\
          D &     13   &   0 &   0 &  13     \\
        H,S &     15   &   0 &  14 &  29      \\
        M,S &     30   &   0 &  25 &  55      \\
        H,D &      8   &   0 &   3 &  11     \\
        M,D &     15   &   0 &  10 &  25      \\
\hline%\\
      total &    246   & 239 & 232 & 717      \\
%\hline
\end{tabular}
\end{center}
\end{table}

Please notice the following changes: Suspicious and dangerous products
encountered in good shops were totally dropped from the statistics
(their existence was not revealed to the public). Suspicious and dangerous 
products from shops with unclear classification (good/bad shops) were 
declared to come from bad shops. Products from good shops which obtained both 
 the  label high quality and dangerous were simply moved into the category 
high quality products (the bad smelt was just concealed) etc.  This is 
frequently the sense in which our beliefs have impact on our attitude towards 
real facts and we will see below that the Dempster-Shafer Theory 
reflects such a view of beliefs. %\\
\end{Bsp}

Let us now define the following function:

\begin{df}
$$Bel_P ^{M}(A)=\Prob{O}{P}(\lnot M(O,\Xi-A))$$
which is the probability that the test M, while being true for A, rejects 
every hypothesis of the form X=$v_i$ for every  $v_i$  not  in  A 
for the population P.
We shall call this function "the belief exactly in the the result 
of measurement". 
\end{df}

Let us define also the function:

\begin{df}
$$Pl_P ^{M}(A)=\Prob{O}{P}(  M(O,A))$$
which  is  the  probability  of  the  test  M  holding  for  
 A for the population P. Let us refer to this function as  the 
"Plausibility of taking any value from the set A".
\end{df}

Last not least be defined the function:
\begin{df}
$$m_P ^{M}(A)=\Prob{O}{P}( \bigwedge_{B;B=\{v_i\}\subseteq A} M(O,B)
  \land \bigwedge_{B;B=\{v_i\}\subseteq \Xi-A} \lnot M(O,B) ) $$
which is the probability that all the  tests  for  the  singleton 
subsets of A are true and those outside of A are  false  for  the 
population P.
\end{df}

Let us illustrate the above concepts with Citizen Coot example:

\begin{Bsp} \label{nonlabelEx}
For the belief function for sold-bottles-population  and the measurement 
function $M ^3$, if we identify probability with relative frequency, we have 
the focal points given in the Table \ref{nonlabtable}:%\\
\end{Bsp}

\begin{table} \label{nonlabtable} 
\caption{Mass and Belief Function under Measurement Method $M ^3$}
\begin{center}
\begin{tabular}{|l|r|r|r|} \hline
Set                           &$m_P ^{M ^3}$& $Bel_P ^{M ^3}$\\
\hline
\{(H,B)                    \} &  20/723     &   20/723       \\
\{(H,G)                    \} & 100/723     &  100/723       \\
\{(H,B),(H,G)              \} &  70/723     &  190/723       \\
\{(M,B)                    \} &  80/723     &   80/723       \\
\{(M,G)                    \} & 100/723     &  100/723       \\
\{(M,B),(M,G)              \} & 110/723     &  290/723       \\
\{(S,B)                    \} &  50/723     &   50/723       \\
\{(S,G)                    \} &   5/723     &    5/723       \\
\{(S,B),(S,G)              \} &  15/723     &   70/723       \\
\{(D,B)                    \} &  10/723     &   10/723       \\
\{(D,G)                    \} &   1/723     &    1/723       \\
\{(D,B),(D,G)              \} &   3/723     &   14/723       \\
\{(H,B),(S,B)              \} &  15/723     &   85/723       \\
\{(H,G),(S,G)              \} &  10/723     &  115/723       \\
\{(H,B),(S,B),(H,G),(S,G)  \} &  14/723     &  299/723       \\
\{(M,B),(S,B)              \} &  30/723     &  160/723       \\
\{(M,G),(S,G)              \} &  20/723     &  125/723       \\
\{(M,B),(S,B),(M,G),(S,G)  \} &  25/723     &  435/723       \\
\{(H,B),(D,B)              \} &   8/723     &   38/723       \\
\{(H,G),(D,G)              \} &   2/723     &  103/723       \\
\{(H,B),(D,B),(H,G),(D,G)  \} &   3/723     &  217/723       \\
\{(M,B),(D,B)              \} &  15/723     &  105/723       \\
\{(M,G),(D,G)              \} &   7/723     &  108/723       \\
\{(M,B),(D,B),(M,G),(D,G)  \} &  10/723     &  336/723       \\
\hline 
 \end{tabular}
\end{center}
\end{table}

It is easily seen that:

% Twierdzenia

\input{mtetheorems.tex}
% Dyskusja i wnioski ko\'{n}cowe
%\input{mteende.tex}

\section{Summary of the New Interpretation}

     The following results have been established in this Section:
\begin{itemize}
\item  concepts of measurement and modified measurement methods  have 
been introduced
\item a concept of labelled population has been developed
\item  it has been shown that a labelled population with the modified 
measurement method can be considered as   Joint  Belief 
Distribution in the sense of MTE,
\item  the process of "relabeling" of a labelled population has been 
defined and shown to be describable as a Belief Distribution.
\item  it has been shown that the  relationship  between  the  Belief 
Distributions of the resulting relabeled population,  the  basic 
population and the relabeling process can be expressed in terms 
of the Dempster-Rule-of-Independent-Evidence-Combination. 
\end{itemize}

     This  last  result  can  be  considered  as  of   particular 
practical importance. The interpretation schemata of MTE
elaborated by 
other authors  (see the remark of Smets below)
suffered from one basic shortcoming: 
if we interpreted population data as well as evidence in terms of 
their MTE schemes, and then combine the evidence  with  population 
data (understood as a Dempster type  of  conditioning)  then  the 
resulting belief function cannot be interpreted in terms  of  the 
population data scheme,  with  subsequent  updating  of  evidence 
making thinks worse till even the weakest  relation  between  the 
belief function and the (selected sub)population is lost.%\\

In this paper we achieve a  break-through:  data  have  the  same 
interpretation scheme after any number of evidential updating and 
hence the belief function can be verified against the data at any 
moment of MTE evidential reasoning.%\\

Properties of the generalized labeling process  
should 
be considered from a philosophical point of view. If we take one by one the 
objects of our domain, possibly labelled previously by an expert in the past,
 and assign a label independently of the actual value of 
the attribute of the object, then we cannot claim in any way that such a 
process may be attributed to the opinion of the expert. Opinions of two 
experts may be independent of one another, but they cannot be independent
of the subject under consideration. This is the point of view with which most 
people would agree, and should the opinions of the experts not depend
on the subject, then at least one of them may be considered as not expert.%\\

This is exactly what we want to point at with our interpretation: 
the precise pinpointing at what kind of independence is assumed 
within the Dempster-Shafer theory is essential for its usability.
Under our interpretation, the independence relies in trying to 
select  a label for fitting to 
an object independently of whatever properties this object has (including its 
previous labeling). The distribution of labels for fitting is exactly 
identical from object to object. The point, where the dependence of object's 
labeling on its properties comes to appearance, is  when the measurement 
method 
states that the label does not fit. Then the object is discarded. From 
philosophical point of view it means exactly that we try to impose our 
 philosophy of life onto the facts: cumbersome facts are neglected and 
ignored. We suspect that this is exactly the justification of the name "belief
function". It expresses not what we see but what we would like to see.\\
Our suspicion is strongly supported by the quite recent statement of Smets 
that  "Far too often, authors concentrate on the static component (how
beliefs are 
 allocated?) and discover many relations between TBM (transferable belief 
model of Smets) 
 and ULP (upper lower probability) models, inner and outer measures 
(Fagin and Halpern \cite{Fagin:89}), random sets (Nguyen \cite{Nguyen:78}), 
probabilities of provability 
 (Pearl \cite{Pearl:90}), probabilities of necessity (Ruspini 
\cite{Ruspini:86}) etc. But these authors 
usually do not explain or justify the dynamic component (how are beliefs 
updated?), that  is, how updating (conditioning) is to be handled (except in 
some cases by defining conditioning as a special case of combination). So I 
 (that is Smets) feel that these partial comparisons are incomplete, 
especially 
as all these interpretations lead to different updating rules." 
(\cite{Smets:92}, pp. 324-325).

Our interpretation explains both the static and dynamic  component 
of the MTE, 
and does not lead to any other but to the Dempster Rule of Combination, hence 
 may be acceptable from the rigorous point of view of Smets. As in the light 
of Smets' paper \cite{Smets:92} we  have presented the only correct 
probabilistic interpretation of the Methematical Theory of Evidence
so far, we feel
to be authorized to claim that our philosophical assessment of the MTE is 
the correct one. %\\

\baselineskip=0.9\baselineskip

\newcommand{\LitStelle}[2]{\bibitem{#1}}

\newcommand{\ReadingsIn}{G. Shafer, J. Pearl eds: Readings in Uncertain 
Reasoning, (ISBN 1-55860-125-2, 
Morgan Kaufmann Publishers Inc., San Mateo, California, 1990)}

\end{document}

%% file: mtetheorems.tex
\begin{th} %THEOREM 3: 
$m_P ^{M}$ is the mass Function in the sense of 
MTE.
\end{th}
\AnfBeweis We shall recall the definition and construction of the DNF 
(Disjunctive Normal Form). If, given an object O of a population P under the 
measurement method M, we look 
at the expression%\\
 $$expr(A)= \bigwedge_{B;B=\{v_i\}\subseteq A} M(O,B)
  \land \bigwedge_{B;B=\{v_i\}\subseteq \Xi-A} \lnot M(O,B)$$
for two different sets $A_1,A_2 \subseteq \Xi$ then clearly 
$expr(A_1) \land expr(A_2)$ is never true - the truth of the one excludes the 
truth of the other. They represent mutually exclusive events in the sense of 
the probability theory. On the other hand:
$$ \bigvee_{A;A \subseteq \Xi} expr(A) = TRUE $$ %\\
hence:
$$ (\Prob{O}{P}(\bigvee_{A;A \subseteq \Xi} expr(A))) = (\Prob{O}{P} TRUE)=1 
$$ %\\
and  due to mutual exclusiveness:\\
$$ \sum_{A;A \subseteq \Xi} (\Prob{O}{P} expr(A)) = 1 $$ %\\
which means:\\
$$ \sum_{A;A \subseteq \Xi} m_P ^M(A) = 1 $$ %\\
Hence the first condition of \rDef{MDef} is satisfied.%\\
Due to the second condition of \rDef{MDef} we have\\
$$(\Prob{O}{P} expr(\emptyset))=1-(\Prob{O}{P}(M(O,
\Xi)))=$$ $$=1-(\Prob{O}{P} TRUE)=1-1=0$$%\\
Hence $$ m_P ^M(\emptyset)=0$$.%\\
 The last condition  is 
satisfied due to the very nature of probability: Probability is never 
negative. So we can  state that $m_P ^M$ is really a Mass Function in the 
sense 
of the MTE.%\\
%\\
\EndBeweis

\begin{th} %THEOREM 1: 
$Bel_P ^{M}$ is a Belief Function in the sense of 
MTE  corresponding  to the  $m_P ^{M}$. 
\end{th}
\AnfBeweis 
Let A be a non-empty set.
By definition %\\
$$M(O,\Xi-A)= \bigvee_{C= \{v_i\} \subseteq \Xi-A} M(O,C)$$%\\
hence by de-Morgan-law:\\
$$\lnot M(O,\Xi-A)= \bigwedge_{C= \{v_i\} \subseteq \Xi-A}\lnot M(O,C)$$%\\
On the other hand, $\lnot M(O,\Xi-A)$ implies $M(O,A)$.\\
But :%\\
$$M(O,A)=
\bigvee_{B \subseteq A} 
\left( \bigwedge_{C;C=\{v_i\}\subseteq B} M(O,C)
  \land \bigwedge_{C;C=\{v_i\}\subseteq A-B} \lnot M(O,C)
\right) $$
\\
So .%\\

\noindent 
$$\lnot M(O,\Xi-A)= \lnot M(O,\Xi-A) \land M(O,A)=$$
$$=
 \bigwedge_{C;C=\{v_i\}\subseteq \Xi-A} \lnot M(O,C)
  \land M(O,A)=
$$
$$ =
 \bigwedge_{C;C=\{v_i\}\subseteq \Xi-A} \lnot M(O,C)
  \land
\left( 
\bigvee_{B \subseteq A} 
\left( \bigwedge_{C;C=\{v_i\}\subseteq B} M(O,C)   \land \right.  \right.
$$
$$
\left. \left. \land  \bigwedge_{C;C=\{v_i\}\subseteq A-B} \lnot M(O,C)
\right)  \right)
=
$$%\\

\noindent 
$$=\bigvee_{B \subseteq A} 
\left(
 \bigwedge_{C;C=\{v_i\}\subseteq B} M(O,C)
  \land \bigwedge_{C;C=\{v_i\}\subseteq \Xi-A} \lnot M(O,C)  
\land \right.$$
$$  \left.  \land \bigwedge_{C;C=\{v_i\}\subseteq A-B} \lnot M(O,C)
\right)
=$$%\\

\noindent 
$$=\bigvee_{B \subseteq A} 
\left( \bigwedge_{C;C=\{v_i\}\subseteq B} M(O,C)  \linebreak 
  \land \bigwedge_{C;C=\{v_i\}\subseteq \Xi-B} \lnot M(O,C) 
\right)
$$%\\
Hence \\
$$\lnot M(O,\Xi-A) =
\bigvee_{B \subseteq A} expr(B)$$%\\
and therefore:%\\
$$(\Prob{O}{P} \lnot M(O,\Xi-A) ) =
(\Prob{O}{P} \bigvee_{B \subseteq A} expr(B))$$%\\
expr(A) being defined as in the previous proof.
As we have shown in the proof of the previous theorem, expressions under the 
probabilities
of the right hand side are exclusive events, and therefore:%\\
$$(\Prob{O}{P} \lnot M(O,\Xi-A) )=\sum_{B \subseteq A} (\Prob{O}{P} 
expr(B))$$ that is:%\\
 $$Bel_P ^M(A \in 2^\Xi ) = \sum_{B \subseteq A} m_P ^M(B)$$%\\
As the previous theorem shows that $m_P ^{M}$ is a MTE Mass 
Function, it suffices to show the above.
%\\
 \EndBeweis

\begin{th} %THEOREM 2: 
$Pl_P ^{M}$ is a Plausibility Function in the sense of 
MTE and it is the Plausibility  Function  corresponding  to 
the  $Bel_P ^{M}$. 
\end{th} %
 \AnfBeweis  By definition:
$$Pl_P ^{M}(A)=\Prob{O}  M(O,A)$$
hence
$$Pl_P ^{M}(A)=1-(\Prob{O}{P} \lnot M(O,A))$$
But by definition:
$$(\Prob{O}{P} \lnot M(O,A))=(\Prob{O}{P} \lnot M(O,\Xi-(\Xi-A)))=Bel_P 
^{M}(\Xi-A)$$ hence 
$$Pl_P ^{M}(A)=1-Bel_P ^{M}(\Xi-A)$$
 \EndBeweis

Two  important remarks must be made concerning this particular 
interpretation:%\\
\begin{itemize}%\\
\item 
Bel and Pl are both defined, contrary to many traditional approaches, as THE 
probabilities and NOT as lower or upper bounds to any probability.%\\
\item
It is Pl(A) (and not Bel(A) as assumed traditionally) that expresses the 
probability of A, and Bel(A) refers to the probability of the complementary 
set $A ^C$.\\
\end{itemize}%\\

Of course, a complementary measurement function is conceivable to revert the 
latter effect, but the intuition behind such a measurement needs some 
elaboration. We shall not discuss this issue in this paper.\\

Let us  also  define  the  following  functions  referred  to  as 
labelled Belief, labelled Plausibility and labelled Mass 
Functions respectively for the labeled population P:

\begin{df}
Let P be a population and $l$ its labeling. Then 

$$Bel_P    ^{M_l}(A)=\Prob{\omega}{P} \lnot M_l(\omega,\Xi-A)$$

$$Pl_P ^{M_l}(A)=\Prob{\omega}{P} M_l(\omega,A)$$

$$m_P ^{M_l}(A)=\Prob{\omega}{P} (\bigwedge_{B;B=\{v_i\}\subseteq A}
 M_l(\omega,B)
  \land \bigwedge_{B;B=\{v_i\}\subseteq \Xi-A} \lnot
 M_l(\omega,B))$$
\end{df}

Let us illustrate the above concepts with Citizen Coot example:

\begin{Bsp} \label{labelEx}
For the belief function for sold-bottles-population P  and the measurement 
function $M ^3$, let us assume the following labeling:\\
$l(\omega)=$\{(H,G),(H,B),(M,G),(M,B),(S,B),(D,B)\}\\
 for every $\omega \in \Omega$, which means that we are convinced that only 
high and moderate quality products are sold in good shops.%.\\
For the population P' under this labeling,
if we identify probability with relative frequency, we have 
the focal points given in the Table \ref{labtable}:%\\
\end{Bsp}

\begin{table} \label{labtable} 
\caption{Mass and Belief Function under 
Modified
Measurement Method $M_{l} ^3$ }
\begin{center}
\begin{tabular}{|l|r|r|}
\hline
Set                           & $m_{P'} ^{M_{l} ^3}$ 
                              &  $Bel_{P'} ^{M_{l} ^3}$ \\
\hline
\{(H,B)                    \} &  20/717     &   20/717       \\
\{(H,G)                    \} & 112/717     &  112/717       \\
\{(H,B),(H,G)              \} &  70/717     &  202/717       \\
\{(M,B)                    \} &  80/717     &   80/717       \\
\{(M,G)                    \} & 127/717     &  127/717       \\
\{(M,B),(M,G)              \} & 110/717     &  317/717       \\
\{(S,B)                    \} &  65/717     &   65/717       \\
\{(D,B)                    \} &  13/717     &   13/717       \\
\{(H,B),(S,B)              \} &  15/717     &  100/717       \\
\{(H,B),(S,B),(H,G)        \} &  14/717     &  184/717       \\
\{(M,B),(S,B)              \} &  30/717     &  175/717       \\
\{(M,B),(S,B),(M,G)        \} &  25/717     &  387/717       \\
\{(H,B),(D,B)              \} &   8/717     &   41/717       \\
\{(H,B),(D,B),(H,G)        \} &   3/717     &  114/717       \\
\{(M,B),(D,B)              \} &  15/717     &  108/717       \\
\{(M,B),(D,B),(M,G)        \} &  10/717     &  228/717       \\
\hline
\end{tabular}
\end{center}
\end{table}

It is easily seen that:

\begin{th} %THEOREM 6: 
$m_P ^{M_l}$ is the mass Function in the sense of 
MTE.
\end{th}
\AnfBeweis To show this is suffices to show that the modified measurement 
method $M_l$ possesses the same properties as the measurement method $M$.\\
Let us consider a labeling $l$ and a population P under this labeling.\\
Let O be an object and L its label under labeling $l$ ($L=l(O)$).
Always  $M_l(O,\Xi)=TRUE$ because by definition $M_l(O,\Xi)=
M(O,\Xi \cap L)=M(O,L)$ and by definition of a labeled population for the 
object's O label L $M(O,L)=TRUE$.\\
Second, the superset consistency is satisfied, because if $A \subset B$ then
if  $M_l(O,A)=TRUE$ then also $M_l(O,A)=
M(O,A \cap L)=TRUE$, but because $A \cap L \subseteq B \cap L$ then also
$M(O,B \cap L)=TRUE$, but by definition $M(O,B \cap L)=M_l(O,B)=TRUE$ and 
thus it was shown that  $M_l(O,A)=TRUE$ implies  $M_l(O,B)=TRUE$ for 
any superset B of the set A.\\
Finally, also the subset consistency holds, because if 
 $M(O,L \cap A)=TRUE$ then there exists a proper subset B of $L \cap A$ such 
that  $M(O,B)=TRUE$. But in this case $B = L \cap B$ so we can formally write:
  $M(O,L \cap B)=TRUE$. Hence we see that $M_l(O,A)=TRUE$ implies the 
existence of a proper subset B of the set A such that $M_l(O,B)=TRUE$.
Hence considering analogies between definitions of $m_P ^M$ and $m_P {^M_l}$ 
as well 
as between the respective Theorems we see immediately that this Theorem  is 
valid.\\
\EndBeweis

\begin{th} %THEOREM 4: 
$Bel_P ^{M_l}$ is a Belief Function in the sense of 
MTE  corresponding  to the  $m_P ^{M_l}$. 
\end{th}
\AnfBeweis
As $M_l$ is shown to be a MTE Mass Function and considering 
 analogies between 
definitions of  $Bel_P ^M$ and $Bel_P {^M_l}$
as well as 
between the respective Theorems we see immediately that this Theorem  is 
valid.
\EndBeweis

\begin{th} %THEOREM 5:
 $Pl_P ^{M_l}$ is a Plausibility Function in the sense of 
MTE and it is the Plausibility  Function  corresponding  to 
the  $Bel_P ^{M_l}$. 
\end{th}
\AnfBeweis
As $M_l$ is shown to be a MTE Mass Function and considering 
 analogies between 
definitions of  $Pl_P ^M$ and $Pl_P {^M_l}$
as well as 
between the respective Theorems we see immediately that this Theorem  is 
valid.
\EndBeweis

This does not complete the interpretation.  

Let  us  now  assume  we  run  a  "(re-)labelling  process"   on   the 
(pre-labelled or unlabeled)
population P. 

\begin{df}
Let $M$ be a measurement method, $l$ be a labeling under this measurement
method, and P be a population under this labeling (Note that the population
may also be unlabeled).
The  {\em (simple) labelling  process}   on    
the
population P 
is defined as a functional 
$LP: 2^\Xi \times \Gamma \rightarrow \Gamma$, where $\Gamma$ is the set of  
all  possible labelings under $M$, 
such that for the given labeling $l$ and a given nonempty
set of attribute values $L$ ($L  \subseteq \Xi$), 
it delivers a new labeling $l'$ ($l'=LP(L,l)$) such that for every object
$\omega \in \Omega$: 

1. if  $M_l(\omega,L)=FALSE$ then  
$l'(\omega)=\emptyset$\\
(that is l' discards a
labeled 
 object $(\omega,l(\omega))$ if $M_l(\omega,L )=FALSE$ 

2. otherwise $l'(\omega)=l(\omega) \cap L $
(that is l' labels the object with $l(\omega) \cap L $ otherwise.
\end{df}

Remark: It is immediately obvious, that the population obtained as the sample 
fulfills the requirements of the definition of a labeled population.%\\

The labeling process clearly induces from P another population P' (a 
population under the labeling $l'$) being a subset of P (hence perhaps 
"smaller" 
than P)   labelled  a 
bit differently. If we  retain  the  primary  measurement 
method M then a  new  modified  measurement  method 
$M_{l'}$ is induced by the new labeling. 

\begin{df} "labelling  process  function" 
$m ^{LP;L }: 2 ^\Xi \rightarrow [0,1]$:
 is defined as:
 $$m ^{LP;L }(L )=1$$  
$$\forall_{B;  B  \in  2^\Xi,B \ne L } m ^{LP;L }(B)=0$$
\end{df}

It is immediately obvious that:

\begin{th} %THEOREM 7.
 $m ^{LP;L }$ is a Mass Function in sense of MTE.
\end{th}

Let  $Bel  ^{LP,L }$  be  the  belief  and  $Pl  ^{LP,L }$  be  the 
Plausibility corresponding to $m ^{LP,L }$. Now let  us  pose  the 
question: what is the relationship between $Bel_{P'} ^{M_{l'}}$, 
 $Bel_P ^{M_l}$,  and $Bel ^{LP,L }$. 

\begin{th} %THEOREM 8: 
\label{thSimpleLab}
Let $M$ be a measurement function, $l$ a labeling, P a population under
this labeling. Let $L $ be a subset of $\Xi$. 
Let $LP$ be a labeling process and let $l'=LP(L ,l)$.
Let P' be a population under the labeling $l'$.
Then 
 $Bel_{P'} ^{M_{l'}}$ is a  combination  via  Dempster's  Combination 
rule of  $Bel ^{M_l}$,  and $Bel ^{LP;L }$., that is:
$$Bel_{P'} ^{M_{l'}} = Bel_P ^{M_l} \oplus Bel ^{LP;L }$$.
\end{th}
\AnfBeweis
Let us consider a labeled object $(O_j,L_j)$ from the population P (before 
re-labeling, that is $L_j=l(O_j)$) which passed the relabeling and became
$(O_j,L_j \cap L )$, that is $L_j \cap L = l'(O_j)$..
Let us define $expr_B$ (before relabeling) and $expr_A$ (after labeling)
as:
$$expr_B((O_j,L_j),A) =  \bigwedge_{B;B=\{v_i\}\subseteq A} M_l(O,B)
  \land$$
$$ \land
 \bigwedge_{B;B=\{v_i\}\subseteq \Xi-A} \lnot  M_l(O,B)$$
and 
$$expr_A((O_j,L_j),A) =  \bigwedge_{B;B=\{v_i\}\subseteq A} M_{l'}(O
 ,B) \land$$
$$  \land \bigwedge_{B;B=\{v_i\}\subseteq \Xi-A} \lnot  M_{l'}(O ,B)$$
Let $expr_B((O_j,L_j),C)=TRUE$ and $expr_A((O_j,L_j),D)=TRUE$ for some C and 
some D. Obviously then for no other C and no other D the respective 
expressions are valid. It holds also that:
$$expr_B((O_j,L_j),C) =  \bigwedge_{B;B=\{v_i\}\subseteq C} M(O_j,L_j \cap B)
\land $$
$$  \land \bigwedge_{B;B=\{v_i\}\subseteq \Xi-D} \lnot  M((O_j,L_j \cap B)$$
and 
$$expr_A((O_j,L_j),D) =  \bigwedge_{B;B=\{v_i\}\subseteq D} M(O_j,L_j \cap 
L 
 \cap B)
  \land$$
$$\land
 \bigwedge_{B;B=\{v_i\}\subseteq \Xi-D} \lnot  M(O_j,L_j \cap 
L  \cap B)$$
In order to get truth on the first expression, C must be a subset of $L_j$, 
and for the second we need D to be a subset of $L_j \cap L $.
Furthermore, for a singleton $F \subseteq \Xi$ 
either $M(O_j,L_j \cap F)=TRUE, M(O_j,L_j \cap L   \cap F)=TRUE$, and then
it belongs to C, $L  $ and D, or%\\
  $M(O_j,L_j \cap F)=TRUE, M(O_j,L_j \cap L   \cap F)=FALSE$, and then
it belongs to C, but not to $L  $ and hence not to D, or%\\
  $M(O_j,L_j \cap F)=FALSE$, so due to superset consistency also
 $M(O_j,L_j \cap L   \cap F)=FALSE$, and then
it belongs  neither to C nor to D
(though membership in $L $ does not need to be excluded).
So we can state that  $D = C \cap L $,
%\\

So the absolute expected frequency of objects for which $expr_A(D)$ holds, is 
given by:%\\
$$ \sum_{C; D= C \cap L } samplecardinality \cdot  m_P ^{M_l}(C)$$%\\
that is:\\
$$ \sum_{C; D= C \cap L } samplecardinality \cdot  m_P ^{M_l}(C)\cdot 
m ^{LP;L }(L)$$%\\
which can be easily re-expressed as: \\
$$ \sum_{C,G; D= C \cap G} samplecardinality \cdot  m_P ^{M_l}(C)\cdot 
m ^{LP;L }(G)$$
%\\
So generally:\\
$$ m_{P'}^{M_{l'}}(D) =c\cdot  \sum_{C,G; D= C \cap G}
  m_P ^{M_l}(C)\cdot m ^{LP;L }(G)$$
with c - normalizing constant.%\\
\EndBeweis

\begin{Bsp} To continue Citizen Coot example let us recall
the function $Bel_P ^{M}$ from Example \ref{nonlabelEx} which is one of 
an unlabeled population.
Let us define the label $$L=
\{(H,G),(H,B),(M,G),(M,B),(S,B),(D,B)\}$$%\\
 as in Example \ref{labelEx}. Let us define the labeling process function as 
 $$m ^{LP;L }(L )=1$$  
$$\forall_{B;  B  \in  2^\Xi,B \ne L } m ^{LP;L }(B)=0$$.
Let us consider the function  $Bel_{P'} ^{M_l}$ from Example \ref{labelEx}.
It is easily seen that:
$$Bel_{P'} ^{M_l}=Bel_{P}^{M} \oplus  Bel ^{LP;L }$$
%
%\\
\end{Bsp}

Let us try  another  experiment, with a more general (re-)labeling process. 
  Instead  of  a  single  set  of 
attribute  values  let  us  take  a  set  of  sets  of  attribute 
values $L ^1, L ^2, ...,L ^k$  (not  necessarily  disjoint)  and 
assign to each one a probability 
$m ^{LP, L ^1, L ^2, ...,L ^k}(A_i)$
of selection.

\begin{df}
Let $M$ be a measurement method, $l$ be a labeling under this measurement
method, and P be a population under this labeling (Note that the population
may also be unlabeled).
Let  us  take  a  set  of (not  necessarily  disjoint) nonempty sets  of  
attribute values $\{L ^1, L ^2, ...,L ^k\}$    and 
let us define the  probability of selection as a function
$m ^{LP, L ^1, L ^2, ...,L ^k}: 2 ^\Xi \rightarrow [0,1]$ such that
$$\sum_{A;A \subseteq \Xi}m ^{LP, L ^1, L ^2, ...,L ^k}(A)=1$$
$$\forall_{A; A \in \{ L ^1, L ^2, ...,L ^k\}} 
m ^{LP, L ^1, L ^2, ...,L ^k}(A)>0$$
$$\forall_{A; A \not\in \{ L ^1, L ^2, ...,L ^k\}} 
m ^{LP, L ^1, L ^2, ...,L ^k}(A)=0$$
 The  {\em (general) labelling  process}   on    
the
population P 
is defined as a (randomized) functional 
$LP: 2^{2^\Xi} \times \Delta
\times  \Gamma \rightarrow \Gamma$, where $\Gamma$ is the set 
of all  possible labelings under $M$, and $\Delta$ is 
a set of all possible probability of selection functions,
such that for the given labeling $l$ and a given 
 set  of (not  necessarily  disjoint) nonempty sets  of  
attribute values $\{L ^1, L ^2, ...,L ^k\}$    and 
a given probability of selection 
$m ^{LP, L ^1, L ^2, ...,L ^k}$
it delivers a new labeling $l"$ such that for every object
$\omega \in \Omega$:

1. a label L, element of the set $\{ L ^1, L ^2, ...,L ^k\}$ 
is sampled randomly according to the probability distribution 
$m ^{LP, L ^1, L ^2, ...,L ^k}$;
This sampling is done independently for each individual object,

2. if  $M_l(\omega,L)=FALSE$ then  
$l"(\omega)=\emptyset$\\
(that is l" discards an object $(\omega,l(\omega))$ if 
$M_l(\omega,L )=FALSE$ 

3. otherwise $l"(\omega)=l(\omega) \cap L $
(that is l" labels the object with $l(\omega) \cap L $ otherwise.)
\end{df}

Again we obtain another ("smaller") population P" under the labeling $l"$  
labelled 
 a bit differently. Also a  new  modified  measurement  method 
$M_{l"}$ is induced by the "re-labelled" population.  
Please notice, that $l"$ is not derived deterministicly. 
Another run of the general (re-)labeling process LP may result in a different
final labeling of the population and hence a different subpopulation under 
this new labeling.%\\

Clearly:

\begin{th} %THEOREM 9. 
$m ^{LP,L ^1,...,L ^k}$ is a Mass Function in sense of MTE.
\end{th}

Let   $Bel   ^{LP;L ^1,...,L ^k}$   be   the   belief    and    $Pl 
^{LP,L ^1,...,L ^k}$  be  the 
Plausibility corresponding to $m ^{LP,L ^1,...,L ^k}$. Now let  us  pose  the 
question: what is the relationship between $Bel_{P"} ^{M_{l"}}$, 
 $Bel_P ^{M_l}$,  and $Bel ^{LP,L ^1,...,L ^k}$. 

\begin{th} %THEOREM 10:
Let $M$ be a measurement function, $l$ a labeling, P a population under
this labeling. 
Let $LP$ be a generalized labeling process and let $l"$
be the result of application of the $LP$ for the set
of labels from the set $\{ L ^1, L ^2, ...,L ^k\}$ 
 sampled randomly according to the probability distribution 
$m ^{LP, L ^1, L ^2, ...,L ^k}$;.
Let P" be a population under the labeling $l"$.
Then 
The expected value 
over the set of all possible resultant labelings $l"$ (and hence
populations P") 
(or, more precisely, value vector) of 
$Bel_{P"} ^{M_{l"}}$ is a  combination  via  Dempster's  Combination 
rule of  $Bel_P ^{M_l}$,  and $Bel ^{LP,L ^1,...,L ^k}$., that is:
$$E(Bel_{P"} ^{M_l'}) = Bel_P ^{M_l} \oplus Bel ^{LP,L ^1,...,L ^k}$$.
\end{th}
\AnfBeweis
By the same reasoning as in the  proof 
of Theorem \ref{thSimpleLab} 
we come to the 
conclusion that for the given label  $L ^i$ and the labeling $l"$ (instead of 
$l'$ %\\
 the absolute expected frequency of objects for which $expr_A(D)$ holds, is 
given by:%\\
$$ \sum_{C; D= C \cap L ^i} samplecardinality \cdot  m_P ^{M_l}(C)
  \cdot  m ^{LP;L ^1,...,L ^k}(L ^i)
$$%\\
as the process of sampling the population runs independently of the sampling 
the set of labels of the labeling process.\\
But  $expr_A(D)$ may hold for any   $L ^i$ such that   $C \subseteq L ^i$, 
hence in all the   $expr_A(D)$ holds for as many objects as:\\
$$ \sum_{i;i=1,...,k} \sum_{C; D= C \cap L ^i} samplecardinality \cdot  m_P 
^{M_l}(C) \cdot m ^{LP;L ^1,...,L ^k}(L ^i)
$$%\\

which can be easily re-expressed as: \\
$$ \sum_{C,G; D= C \cap G} samplecardinality \cdot  m_P ^{M_l}(C)\cdot 
m ^{LP;L ^1,...,L ^k}(G)$$
%\\
So generally:\\
$$ E(m_{P"}^{M_{l"}}(D) )=c\cdot  \sum_{C; D= C \cap G}  m_P ^{M_l}(C)\cdot 
m ^{LP;L ^1,...,L ^k}(G)$$
with c - normalizing constant.%\\
Hence the claimed relationship really holds.%\\
 \EndBeweis

\begin{Bsp}
The generalized labeling process and its consequences
may be realized in our Citizen Coot example 
by randomly assigning the sold bottles for evaluation to two
"experts", one of them 
- considering about 30 \% of the bottles - is running the full $M$ test 
procedure, and  the other - having to  consider the remaining 70 \% of 
 checked bottles - makes it easier for himself  by making use of his belief 
in the labeling $l$ of Example \ref{labelEx}.  %\\
\end{Bsp}

% 